\documentclass[
  journal=nlp,
  manuscript=article,
  year=2026,
  volume=??,
]{cup-journal}

\usepackage{import}
\usepackage{orcidlink}
\usepackage{amssymb}
\usepackage{amsfonts}
\usepackage{amsmath}
\usepackage{enumitem}
\usepackage{csquotes}
\usepackage{textcomp}
\usepackage[dvipsnames, table]{xcolor}
\usepackage{multirow}
\usepackage{vcell}
\usepackage{url}
\usepackage{listings}
\usepackage{wrapfig}
\usepackage{rotating}
\usepackage{colortbl}
\usepackage[english]{babel}

\usepackage{hyperref}
\usepackage{cleveref}

\usepackage{algorithmicx}
\usepackage{algorithm}
\usepackage{algpseudocode}

\newcommand{\Step}[1]{\algrenewcommand{\alglinenumber}[1]{Step ##1: } #1}
\newcommand{\NoNumber}{\algrenewcommand{\alglinenumber}[1]{\setcounter{ALG@line}{\numexpr##1-1} \ \ \ \ \ \ \ \ \ \ }}

\newcommand{\multiline}[1]{%
    \begin{tabularx}{\dimexpr\linewidth-\ALG@thistlm}[t]{@{}X@{}}
        #1
    \end{tabularx}
}

\usepackage{amsmath}
\usepackage[nopatch]{microtype}
\usepackage{booktabs}

\title{Repeatability is not recovery: Quantifying algorithmic stability and topic recovery in Latent Dirichlet Allocation}

\author{Saranzaya Magsarjav \orcidlink{0000-0002-8776-3312}}
\email[Saranzaya Magsarjav]{saranzaya.magsarjav@adelaide.edu.au}

\author{Lewis Mitchell \orcidlink{0000-0001-8191-1997}}
\author{Jonathan Tuke \orcidlink{0000-0002-1688-8951}}
\author{Melissa Humphries \orcidlink{0000-0002-0473-7611}}
\affiliation{School of Mathematical Sciences, Adelaide University, Adelaide, SA, Australia}

 %% First letter not capped

\hypersetup{hidelinks}

\begin{document}

\begin{abstract}

% Topic models are often judged by the consistency of their outputs across repeated runs, implicitly assuming that repeatable topic output is a successful recovery of the underlying topics. We show that this assumption is false: repeatability is not recovery. We introduce a stability framework that jointly measures consistency among repeated runs and accuracy relative to known ground truth. Because real-world corpora lack known topic structures, we generate synthetic corpora using the Latent Dirichlet Allocation (LDA) generative process, enabling direct evaluation of topic recovery. Across 50 repeated LDA runs on each corpus, we find that LDA reliably identifies the correct number of topics and frequently converges to highly consistent topic solutions. However, these repeatable solutions frequently fail to recover the true generating topics. Thus, internal stability should not be interpreted as evidence of correctness. Our results illustrate that stability and recovery are distinct properties of topic models and should be evaluated separately. Consequently, topic-model outputs should be validated using multiple complementary criteria before supporting substantive conclusions, particularly in high-stakes applications.

Topic modelling uncovers hidden topics in unlabelled text. As most topic models are stochastic, rerunning a topic model can and does produce different topics. Repeated outputs are often interpreted as evidence that the underlying topics have been recovered. However, repeatable outputs do not necessarily imply accurate recovery of the true topic structure. Here we show that repeatability is not the same as recovery. We introduce an Instability framework that uses the mean and variance of topic-similarity scores to measure consistency across repeated fits and accuracy relative to known ground truth. To provide this ground truth, we generated 50 simulated corpora using the Latent Dirichlet Allocation generative process. Each corpus used the same ground-truth distributions but contained separately sampled documents and was fitted 50 times using fixed random seeds. We used Jensen–Shannon Similarity for complete distributions, Jaccard Similarity for the top 10 terms, and Rank-Biased Overlap for their ordering. We found that LDA generally returned topics that were more similar across repeated fits than they were to the true generating topics. The Instability measures generally provided a signal near \(K=10\), the true topic count, although the strength of this signal depended on the generating distribution and similarity measure. For the Uniform Separable distribution, LDA recovered the highest-ranked terms and their ordering more accurately than the complete topic-word probabilities. NPMI agreed with the Instability measures for this clearly separated distribution but generally favoured smaller topic counts for the overlapping distributions. For the 20 Newsgroups dataset, Jaccard and Rank-Biased Overlap Instability were lowest near the 20 labelled categories, while Jensen–Shannon Instability and NPMI showed different patterns. Overall, these results show that repeatability, recovery, and coherence capture different properties of topic-model performance. Therefore, repeatable topics should not be treated as evidence that the fitted topics are correct, and the measures should be considered together.
\end{abstract}

\section{Introduction}

Topic modelling in Natural Language Processing is a process for uncovering the hidden `topics' in a large unlabelled body of texts \citep{blei2009topic}. Topic modelling is widely used in information retrieval \citep{hambarde2023information}, content summarisation \citep{murakami2017corpus, onah2022data}, trend analysis \citep{lau2012line, corti2022social}, and linguistic analytics \citep{howes2013investigating}. Due to its wide range of applications, topic modelling is used across disciplines, from the humanities to the health and life sciences, and more \citep{vayansky2020review, abdelrazek2023topic}. 

Recently, Large Language Models (LLMs) have been used quite effectively for topic analysis \citep{doi2024topic}. However, sometimes more transparency and insight are required than an off-the-shelf LLM can provide. For example, when using topic analysis for high-risk decision-making, a clear understanding of topic stability and coherence is required to improve decision-making. For this precise reason, topic analysis must still be conducted outside LLMs, and this paper aligns with those purposes.

The diversity in applications and disciplines necessitates robustness in the topic model, particularly in terms of replicability and the accuracy of identifying latent topics. In practice, repeated topic-model outputs are often interpreted as evidence that the underlying latent topics have been successfully recovered. However, repeatable outputs do not necessarily imply accurate recovery of the true latent topic structure. Due to the stochastic nature of some probabilistic topic modelling and inference, including LLMs, re-running the topic models yields different latent topics.

The change in latent topics can occur when the corpus is shuffled, rare words are removed, and documents are removed, etc. Theoretically, these should not have a major effect on the outcome; however, in practice, they can yield very different latent topics. These changes are, in part, due to the random initialisation for inference and modelling, but can also be impacted by the strength of the topics present in the corpus. The variation in the output is considered the `stability' of the model. Existing approaches typically quantify this stability by measuring the consistency of topic-model outputs across repeated runs. While consistency is important, it cannot determine whether the recovered topics correspond to the true underlying topic structure. Variations and instability create problems in analysis, as they lead to inconsistent summarisation, lack of replicability and reliability, poor interpretation, and uncertainties \citep{ballester2022robustness}. If these slight changes cause large output discrepancies, it prompts questioning of what the topic model is capturing. Are the topic models producing `real' coherent topics, or are they modelling noise? If there are no strong topics present in the corpus, how can we determine what it is modelling? To address this problem, we propose a new instability measure that combines consistency across repeated runs with accuracy relative to simulated known ground truth. We applied these measures to a simulated dataset we generated to provide ground truth, allowing us to distinguish repeatable topic solutions from those that accurately recover the underlying topics.

In the literature, a common approach to quantify the stability of topic models is to rerun the topic model multiple times and use clustering to determine the true number of topics using different measures. \citet{mantylaMeasuringLDATopic2018} replicated LDA runs multiple times and used K-medoids to cluster the different topics with Rank-Biased Overlap as the stability measure to assist with determining the number of topics within the corpus. \citet{greeneHowManyTopics2014} and \citet{belfordStabilityTopicModeling2018a} focused on measuring stability using term stability based on Jaccard Similarity. These methods compare the overlap between the top terms of topics rather than directly comparing the topic-word probability distributions. However, these methods return to the problem of a lack of ground truth. To address the lack of labelled data, the dataset is often simulated using the generative processes of the topic models.

\citet{taylorSimLDAToolTopic2023} and \citet{weisserPseudodocumentSimulationComparing2023} created simulation methods to evaluate topic model performance. \citet{taylorSimLDAToolTopic2023} used Kullback-Leibler divergence (KLD) to determine the accuracy of topic models. \citet{weisserPseudodocumentSimulationComparing2023} used document assignment, human evaluations, and matrix correlation to determine the accuracy of different topic models. These methods focus on determining how accurately different topic models recover the underlying topic structure, rather than measuring the consistency of repeated runs of the same model.

We have seen that the two main approaches used to evaluate topic models are measuring consistency across repeated runs or measuring recovery using simulated data. Choosing better metrics can accurately quantify differences; however, without a ground truth to compare against, the quantities are arbitrary, and it becomes harder to interpret the quantification of stability as it becomes more complex. The main concern in quantifying model stability is the lack of high-quality, labelled data to test the results. Labelled data helps determine if the topic models are modelling the topics or a different type of structure in the data. The lack of high-quality data necessitated the creation of simulated data. These simulated data are useful because they provide a ground truth for comparing topic model outputs. However, in most of the literature where simulated data are used, different topic models are compared rather than the consistency of a single model. As a result, it remains unclear whether a model is repeatable because it has recovered the true topics or because it consistently returns similar topic solutions. Therefore, we combine the two approaches by using simulated datasets to determine accuracy and replicated runs to quantify consistency.

Using a simulated corpus as input to evaluate model stability on replicated runs captures not only recovery accuracy but also reproducibility and consistency. Most of the literature focuses on one aspect. By combining these two, we can determine whether a model consistently recovers the true underlying topics or simply returns repeatable topic solutions. This distinction allows us to separate repeatability from recovery.

In this paper, we will focus on the Latent Dirichlet Allocation (LDA) generative process to generate the labelled dataset \citep{jelodar2019latent}. As we control the generation process, different topic-word probability distributions are considered. To assess different aspects of topic model accuracy, we have used the Jaccard Index, Jensen-Shannon Distance, and Rank-Biased Overlap. LDA was run multiple times on the generated data to capture variability in LDA output. Using the similarity measures, we defined a new Instability measure that captures both recovery accuracy and consistency across repeated runs. With this new Instability measure, we show that LDA is internally consistent, \textit{i.e.}, finds similar topics across repeated runs; however, these repeatable topic solutions are not necessarily the `true’ generating topics.

This paper makes two main contributions: data generation and the Instability measure. First, the Instability measure is defined using three different similarity metrics. The second is the data generation process and how the dataset was processed for analysis.

\section{Method}

\subsection{Stability Measurement} \label{sec:stability}

We have defined model stability using both recovery accuracy and consistency across multiple runs; therefore, we need not only the mean accuracy but also the variability in the outputs. To measure this, we compared the true distributions to the output distributions of Latent Dirichlet Allocation (LDA) using different similarity measures (see Section \ref{sec:similarity}). If the output distributions of LDA correspond with the true topic-word distributions, then we should see relatively large values in the similarity measure. High values in the similarity measure indicate that LDA is correctly identifying the specified topics in the corpus. If the LDA outputs are consistent, then the variance of these measures should be low across multiple runs, \textit{i.e.}, the topics found using LDA are similar when run multiple times. Therefore, when comparing the variance against the mean, we want these values to be closer to the point \((1,0)\), as all our similarity measures have a maximum value of 1. We define the \textit{Instability} measure as the Euclidean distance to this point.

Let \(m_i\) denote the set of similarity scores obtained for the \(i\)th corpus across repeated LDA runs, and let \(c_i=(\bar{m}_i,\operatorname{var}(m_i))\) denote their mean and variance. For the recovery comparison, \(m_i\) contains the similarities between each LDA output and the true generating distribution. For the repeated-run comparison, \(m_i\) contains the similarities between the first LDA output and the outputs from the remaining runs.

The Instability measure is defined as the average Euclidean distance between \(c_i\) and the ideal point \((1,0)\) across \(t\) corpora:

\begin{equation}
    \operatorname{Instability}_{t}=
\frac{1}{t}
\sum_{i=1}^{t}
\sqrt{
(\bar{m}_{i}-1)^{2}
+
\operatorname{var}(m_i)^{2}
}.
\label{eq:instability}
\end{equation}

A low Instability value therefore corresponds to a high mean similarity and low variation across repeated runs. When \(m_i\) contains similarities to the true generating distribution, the mean captures recovery accuracy and the variance captures the consistency of recovery across runs. When \(m_i\) contains similarities between repeated LDA outputs, the measure captures internal repeatability only. Because the similarity scores are bounded between 0 and 1, their means are also bounded between 0 and 1 and their variances are bounded between 0 and (1/4). The ideal value of zero is obtained when the mean similarity is one and its variance is zero.

\subsection{Similarity Measures}\label{sec:similarity}

We used three similarity measures to quantify the similarity between the generated topics and the LDA output topics. One compares distributions, and another is for set comparison. The third captures the rank within the set. The distribution similarity measure was used to compare the `true' generated distributions to the found output distributions. The set comparisons were used to compare the top \(n\) terms of the topic. The idea is that the top terms summarise the topic itself. These similarity measures were then used to define the \textit{Instability} measure described in Section~\ref{sec:stability}.

The specific similarity measures used are the Jensen-Shannon Similarity (JSS), the Jaccard Index Similarity (JIS), and Rank-Biased Overlap (RBO).

\subsubsection{Jensen-Shannon Similarity}

Jensen-Shannon Divergence was used to measure the distance between distributions. As the output distribution changes, JSS can detect smaller shifts, thereby capturing nuanced changes. Jensen- Shannon Distance (JSD) \citep{endresNewMetricProbability2003} is a metric that is derived from Kullback-Leibler divergence (KLD), defined as 
\[
    D_{KL}(P|| Q) = \sum_{x \in X} P(x)\log \left(\frac{P(x)}{Q(x)}\right),
\] 

where \(P\) and \(Q\) are the discrete probabilities over \(X\). To avoid continuity problems, the following conventions are used
\[ P(x)\log \frac{P(x)}{Q(x)} = 
    \begin{cases}
        p \log(p/q) \quad & p>0, q>0\\
        0 & p = 0, q\geq 0\\
        \infty & \text{else}.
    \end{cases}
\]

KLD is not a true distance metric as it is asymmetric and does not satisfy the triangle inequality. These are avoided in Jensen-Shannon Divergence by averaging the divergence between the two distributions.

\[
    D_{JS}(P || Q) = \frac{1}{2}D_{KL}(P|| M) + \frac{1}{2}D_{KL}(Q || M), 
\]

where \(M = \frac{1}{2} (P+Q)\). The square root of Jensen-Shannon Divergence, \textit{i.e.}, \(d_{JS}(P||Q) = \sqrt{D_{JS}(P|| Q)}\) is taken to make it a distance metric. This results in a symmetric and always finite distance metric. JS distance is bounded by 0 and 1, given that the logarithm is in base 2, where 0 is indicative of \(P\) and \(Q\) being the same distribution. Therefore, to convert the distance measures to a similarity score, the Jensen-Shannon Distance is subtracted from 1, \textit{i.e.,} 

\begin{equation}\label{eq:jsd}
    JSS(P|| Q) = 1 - d_{js}(P|| Q)
\end{equation}

Using this measure, the difference between the topic-word distributions can be quantified.

\subsubsection{Jaccard Similarity}

Jaccard Similarity Index is commonly used to compare the top words in each topic by treating them as sets. By focusing only on the top words, it enables a straightforward comparison across topics based on word overlap. This is useful because JIS measures the extent of shared words between topics, unaffected by minor changes in word order or weight. However, because it ignores word rank and frequency, it may miss subtler differences.

Jaccard Index measures the similarity of two sets, where a set is the top \(n\) tokens of each topic. It measures the overlap of two finite sets by taking the intersection over the union.

\[
    JI(A, B) = \frac{|A \cap B|}{|A \cup B|}
\]

The Jaccard Index is bounded between 0 and 1, where 0 indicates that the two sets are completely different, and 1 indicates that the two sets are equivalent. If set A is a subset of B, then the Jaccard Index would return a value less than 1. 

\subsubsection{Rank-Biased Overlap}

Rank-Biased Overlap (RBO) measures the similarity between two ranked lists while giving greater importance to agreement near the beginning of the lists \citep{rbo}. Unlike Jaccard Similarity, RBO considers the ranking of the terms rather than treating them as unordered sets. It can also compare lists that do not contain exactly the same terms.

Let \(S\) and \(T\) be the ranked lists of terms from two topics, and let \(S_d\) and \(T_d\) contain the terms up to depth \(d\). The overlap at depth \(d\) is

\[
X_d = |S_d \cap T_d|,
\]

and the agreement at that depth is

\[
A_d=\frac{X_d}{d}.
\]

RBO is defined as a weighted sum of the agreement at each depth:

\[
RBO(S,T;p)
=(1-p)\sum_{d=1}^{\infty}p^{d-1}A_d,
\]

where \(p\) is the persistence parameter. The persistence parameter controls how quickly the weight decreases as the depth increases. Smaller values of \(p\) place more weight on the first few terms, while larger values allow terms further down the rankings to contribute more to the comparison.

For this analysis, we used the 10 terms with the highest topic-word probabilities and set \(p=0.9\). The expected evaluation depth is \(1/(1-p)=10\), so this value corresponds with the use of the 10 highest-ranked terms. As the observed rankings were limited to 10 terms, the extrapolated RBO score was calculated as

\[
RBO(S,T;0.9)
=(1-0.9)\sum_{d=1}^{10}0.9^{d-1}A_d
+0.9^{10}A_{10}.
\]

This extrapolated form accounts for the unobserved portion of the rankings while retaining greater weight on agreement near the top. RBO is bounded between 0 and 1, where 0 indicates no overlap and 1 indicates identical rankings.

The topic-word probabilities were used to determine the order of the terms in each ranked list. Once the rankings were constructed, RBO compared the identities and ranks of the terms.

For both Jaccard Similarity and Rank-Biased Overlap, we used the 10 terms with the highest topic-word probabilities for each topic. For Jaccard Similarity, these terms were treated as an unordered set. For Rank-Biased Overlap, the ordering of the terms was retained.

Using these three different similarity measures, different levels of variation in the LDA outputs are measured.

\subsection{Data Generation}

Latent Dirichlet Allocation (LDA) is a generative probabilistic model of the collection of documents. The main assumption is that each document is a mixture model over the set of latent topics, where each topic is a discrete distribution over the words. LDA is a soft clustering as documents are assigned to multiple topics with associated probabilities. One major assumption is that the number of topics is known and fixed. The LDA process produces two final structures: a document with topic probabilities and a topic with word distributions. As LDA is a generative process (see Algorithm \ref{alg:lda}), we used the algorithm to generate a theoretical corpus. We generated these corpora to uncover an underlying truth, which we then compared with the model's outputs to assess its performance. Comparing the generated distributions with the LDA output provided an accuracy measure for the LDA topic model.

% \begin{algorithm*}[ht!]
%     \caption{Algorithm LDA} \label{alg:lda}
%     \renewcommand{\thealgorithm}{}
%         \begin{algorithmic}[1]
%             \Step \State Choose \(\theta_i \sim Dir(\alpha)\) where \(i \in \{1,2,\hdots, M\}\)
%             \Step \State Choose \(\phi^{(k)} \sim Dir(\beta)\); \(N(.,\sigma^2)\); \(U(.,.)\); where \(k \in \{1,2,\hdots, K\}\)
%             \Step 
%             \For {\(w_i \in d\)}
%             \NoNumber \State {Choose \(z_i \sim Multinomial(\theta_i)\)};
%             \NoNumber \State {Choose \(w_i \sim Multinomial(\phi^{z_i})\)};
%             \EndFor
%         \end{algorithmic}

% \end{algorithm*}

% \begin{itemize}[noitemsep]
%     \item \(Dir(.)\) is the Dirichlet Distribution
%     \item \(N(., \sigma^2)\) is the Normal Distribution
%     \item \(U(.,.)\) is the Uniform distribution 
%     \item \(z_i\) is the latent topic assigned to the word \(w_i\)
%     \item \(\theta\) is the probability of topic \(z\) occurring in document
%     \item \(\phi\) is the probability of word \(w_i\) occurring in topic 
%     \item \(\alpha\) is the prior weight of topic \(k\) in a document
%     \item \(\beta\) is the prior weight of \(w_i\) in topic \(z_i\)
%     \item \(M\) is the number of total documents
%     \item \(d\) is the lengths of each document
%     \item \(V\) is the number of unique tokens/text
%     \item \(K\) is the number of topics
% \end{itemize}

\begin{algorithm*}[ht!]
    \caption{LDA corpus-generation process}
    \label{alg:lda}

    \begin{algorithmic}[1]

        \Step{\State Construct the topic-word distributions.}
        \NoNumber
        \For{\(k = 1,\ldots,K\)}
            \State Construct \(\phi^{(k)}\) using the selected
            Dirichlet, Normal, or Uniform Separable distribution.
            \State Normalise \(\phi^{(k)}\) so that
            \(\sum_{v=1}^{V}\phi^{(k)}_{v}=1\).
        \EndFor

        \Step{\State Generate the documents.}
        \NoNumber
        \For{\(i = 1,\ldots,M\)}
            \State Sample the document length
            \(N_i \sim
            \operatorname{DiscreteUniform}\{100,\ldots,300\}\).
            \State Sample the document-topic distribution
            \(\theta_i \sim \operatorname{Dirichlet}(\alpha)\).

            \Step{\State Generate the terms in document \(i\).}
            \NoNumber
            \For{\(n = 1,\ldots,N_i\)}
                \State Sample the topic assignment
                \(z_{in} \sim
                \operatorname{Categorical}(\theta_i)\).
                \State Sample the term
                \(w_{in} \sim
                \operatorname{Categorical}
                \left(\phi^{(z_{in})}\right)\).
            \EndFor
        \EndFor
    \end{algorithmic}
\end{algorithm*}

\begin{itemize}[noitemsep]
    \item \(\phi^{(k)}\) is the topic-word probability distribution
    for topic \(k\).
    \item \(\theta_i\) is the document-topic probability distribution
    for document \(i\).
    \item \(z_{in}\) is the latent topic assigned to term position \(n\)
    in document \(i\).
    \item \(w_{in}\) is the term generated at position \(n\) in
    document \(i\).
    \item \(\alpha\) is the parameter of the document-topic Dirichlet
    distribution. In the experiments, \(\alpha_k=0.01\) for every
    topic \(k\).
    \item \(\beta\) is the parameter of the topic-word Dirichlet
    distribution when the Dirichlet generating distribution is used.
    \item \(M\) is the total number of documents.
    \item \(N_i\) is the sampled length of document \(i\).
    \item \(V\) is the vocabulary size.
    \item \(K\) is the number of topics.
\end{itemize}

The idea of stability cannot be determined by running the LDA topic model only once, as we need to capture the variability. Therefore, 50 different corpora were generated with the same initial conditions, and each was shuffled and run through the LDA topic model 50 times. In theory, the best-performing measures should indicate that the chosen topic matches the predefined number of topics.

For each document, the document length was sampled from a discrete Uniform distribution between 100 and 300 terms. The document-topic distribution was generated using a symmetric Dirichlet distribution, with \(\alpha_k=0.01\) for every topic \(k\). We then ran through the LDA generative process until the sampled document length was reached.

Because some terms had very low probabilities, they did not always occur during the initial corpus generation. To ensure that all predefined terms were represented, occurrences of the missing terms were inserted by replacing existing terms. For each missing term, the number of replacements was sampled from a Poisson distribution with \(\lambda=3\). If the sampled value was 0, it was replaced with 1 so that every missing term appeared at least once. The topic-word distributions were reweighted for each missing term across the topics and used to assign the term to a topic. The document-topic distributions were then used to select the documents in which the replacements were made. Within the selected documents, the most frequent existing terms were replaced with the missing term. Therefore, the final document lengths remained between 100 and 300 terms. These replacements changed the realised term counts but preserved the sampled document lengths and ensured that all predefined terms were represented. Finally, the terms within each document were shuffled randomly.

The document-generation process was repeated 50 times to produce 50 distinct corpora with the same ground-truth distributions and predefined initial parameters. Different random seeds were used to generate each corpus, so the sampled documents differed even though the underlying ground truth remained the same. Each corpus was then fitted with the LDA topic model 50 times, using a different fixed random seed for each fit. All 50 LDA fits for every corpus were retained, with no fits excluded from the analysis.

We set the number of topics \(K = 10\) to the true number of topics and ran the topic model for \(K = 5\) to \(K = 20\). We tested five topic-word distributions: the Dirichlet Distribution with parameters 0.0001 \(Dir_{small}\) and 0.001 \(Dir_{mid}\), the normal distribution with standard deviations of 1 \(N_{1}\) and 10 \(N_{10}\), and finally, the Uniform Separable distribution \(Uni_{sep}\). 
% These distributions were chosen because they provide a good variety for comparison, and a separable uniform distribution allows us to create a baseline for what should work best in this setting. 

\subsubsection{Construction of the Normal and Uniform Separable distributions}

To generate the Normal topic-word distributions, we divided the terms into equal sections based on the number of topics. Each section corresponded to one topic. For example, with 100 terms and 10 topics, each topic was assigned a section containing 10 terms. The Normal distribution for each topic was centred at the midpoint of its section. The standard deviation was measured in terms of the number of term positions. We used standard deviations of 1 \(N_1\) and 10 \(N_{10}\), where the larger standard deviation produced greater overlap between neighbouring topics. When part of a Normal distribution extended beyond the first or final term in the vocabulary, the probability mass was redistributed to the terms at the opposite edge. The probabilities were then normalised so that the topic-word probabilities for each topic summed to 1.

The Uniform Separable distribution \(Uni_{\mathrm{sep}}\) used the same division of terms into topic sections. However, the distribution was cut off completely at the boundaries of each section. All terms within a section were assigned equal probability, while terms outside the section were assigned probability 0. For illustrative purposes, consider a vocabulary containing 100 terms and 10 topics, where each topic is assigned a section containing 10 terms. The actual experiments used 1,000 unique terms and 10 topics. This produced completely separated topics and provided a baseline for a topic structure that should be easier for LDA to recover.

For analysis in the following sections, the number of documents was set to 1000, and the number of unique terms was set to 1000. For each run through LDA, the similarity measures defined in Section \ref{sec:similarity} were used.

\subsection{Processing Data}

All LDA models were fitted using the tomotopy\footnote{tomotopy version 0.12.4; documentation available at \url{https://bab2min.github.io/tomotopy/}} implementation of LDA, which uses Gibbs sampling for inference. Each model was run for the full 1,000 iterations, with the first 100 iterations used as burn-in. Perplexity was used as a convergence diagnostic rather than as a stopping rule. The absolute change between successive recorded perplexity values was compared with a tolerance of 0.0001, but all model fits completed the full 1,000 iterations regardless of when this threshold was reached. Apart from the number of topics and the fixed random seed for each fit, the LDA hyperparameters were left at the default values provided by tomotopy. All similarity comparisons were based only on the estimated topic-word distributions.

One thing to note is that topic outputs do not have an order; therefore, we need to find the corresponding topics across all runs. The first topic model output was used as the baseline for comparing the remaining runs. From comparisons between the first and the remaining runs, the Jensen-Shannon (JSS) similarities are measured using the topic-word distribution. Comparing all topics creates a matrix of JSS values, with the first entry being the JSS between topic 1 of the first output and topic 1 of the second output. A greedy algorithm was applied to determine the corresponding topics, as preliminary experiments showed that it did not make much difference compared to the Hungarian assignment and was computationally efficient.

\begin{algorithm*}[ht!]     
    \caption{Corpus Comparison}
    \label{alg:corpus}
    \renewcommand{\thealgorithm}{}
        \begin{algorithmic}[1]
            \Step
            \For {\(c \in 1:50\)}\\
                \NoNumber \(D = D_c\)
                \For {\(k \in \mathbf{K}\)}
                \NoNumber \State {\(\phi_1, \theta_1=lda(D, k)\)};
                \NoNumber
                        \If{\(k == k_{true}\)} 
                        \NoNumber \State  {\(JSS_{true}, JIS_{true}, RBO_{true} =\) compare\((\phi_{true}, \phi_1)\)};
                        \EndIf
                \Step
                    \For {\(i\) in 2:50}
                    \NoNumber \State {shuffle(D)}
                    \NoNumber \State {\(\phi_i, \theta_i=lda(D, k)\)};
                    \NoNumber \State {\(JSS_i, JIS_i, RBO_i =\) compare\((\phi_1, \phi_i)\)};
                    \Step
                        \If{\(k == k_{true}\)} 
                        \NoNumber \State  {\(JSS_{true}, JIS_{true}, RBO_{true} =\) compare\((\phi_{true}, \phi_i)\)};
                        \EndIf
                    \EndFor
                \EndFor
            \EndFor
    \end{algorithmic}
\end{algorithm*}

To ensure consistency, the process was repeated across all simulated runs to obtain all matching topics (see Algorithm \ref{alg:corpus}). The similarity measures between the theoretical distributions/terms and the output distributions/terms were only measured when the chosen number of topics was the same as the initial number of topics. Only the true number of topics was compared, as it is not meaningful to compare distributions of different sizes in this context. 

For each simulated corpus \(i\), the topic-word distributions from the first LDA fit, \(\phi_{i1}\), were used as the baseline. The topic-word distributions from each remaining fit, \(\phi_{ij}\), where \(j=2,\ldots,50\), were compared with \(\phi_{i1}\). The topics were first aligned using the matching procedure described above. The similarity values across the matched topics were then averaged, yielding a single similarity score for each repeated-fit comparison. For each corpus, these similarity scores \(m_i\) were used to calculate their mean and variance, thereby yielding the corpus-specific Instability value. This process was repeated for all 50 simulated corpora. Finally, the 50 corpus-specific Instability values were averaged to give the reported \textit{Instability} value.

NPMI topic coherence was calculated using the \texttt{c\_npmi} implementation in \textit{tomotopy} version 0.12.4 \citep{roder2015}. For each fitted model, coherence was calculated using the 10 highest-probability terms from each topic and a sliding reference window of 10 tokens. The topic-level coherence scores were then averaged to yield a single coherence value for each fit. NPMI coherence was calculated over the same range of topic numbers and compared with the Instability measures to determine whether coherence and repeatability provided similar evidence about the appropriate number of topics.

We also tested our stability measure on the well-known 20 Newsgroups dataset \citep{20newsgroups}. The dataset consists of over 18,000 newsgroup posts/messages across 20 topics, ranging from religion to sports. The entire 20 Newsgroups dataset was used, with headers, footers, and quoted text removed before preprocessing. The remaining text was converted to lower case and tokenised using the \texttt{SimpleTokenizer} provided by \textit{tomotopy}. Stop words were removed using the NLTK English stop-word list, and the remaining terms were stemmed using the NLTK Porter stemmer. We applied the same process, running the documents through LDA with 50 different LDA fits using different fixed random seeds, with no fits excluded, over a range of topic numbers \(K=13\) to \(K=28\) to determine stability.

We expect the \textit{Instability} values to be smaller for distributions with greater separation between the topic-word distributions. Therefore, the expectation is that the uniform distributions should have the lowest values, followed by the Dirichlet distribution with the smaller parameter, then the Dirichlet distribution with the larger parameter, then the Normal distribution with standard deviation 1, and finally the Normal distribution with standard deviation 10, which is expected to be the least stable. As one of the assumptions of LDA is that the topic-word distributions are Dirichlet, we also expect the Dirichlet-generated data to produce lower \textit{Instability} values than the Normal distributions.

\section{Results}

\begin{figure}[ht]
    \centering
    \includegraphics[width=0.9 \textwidth]{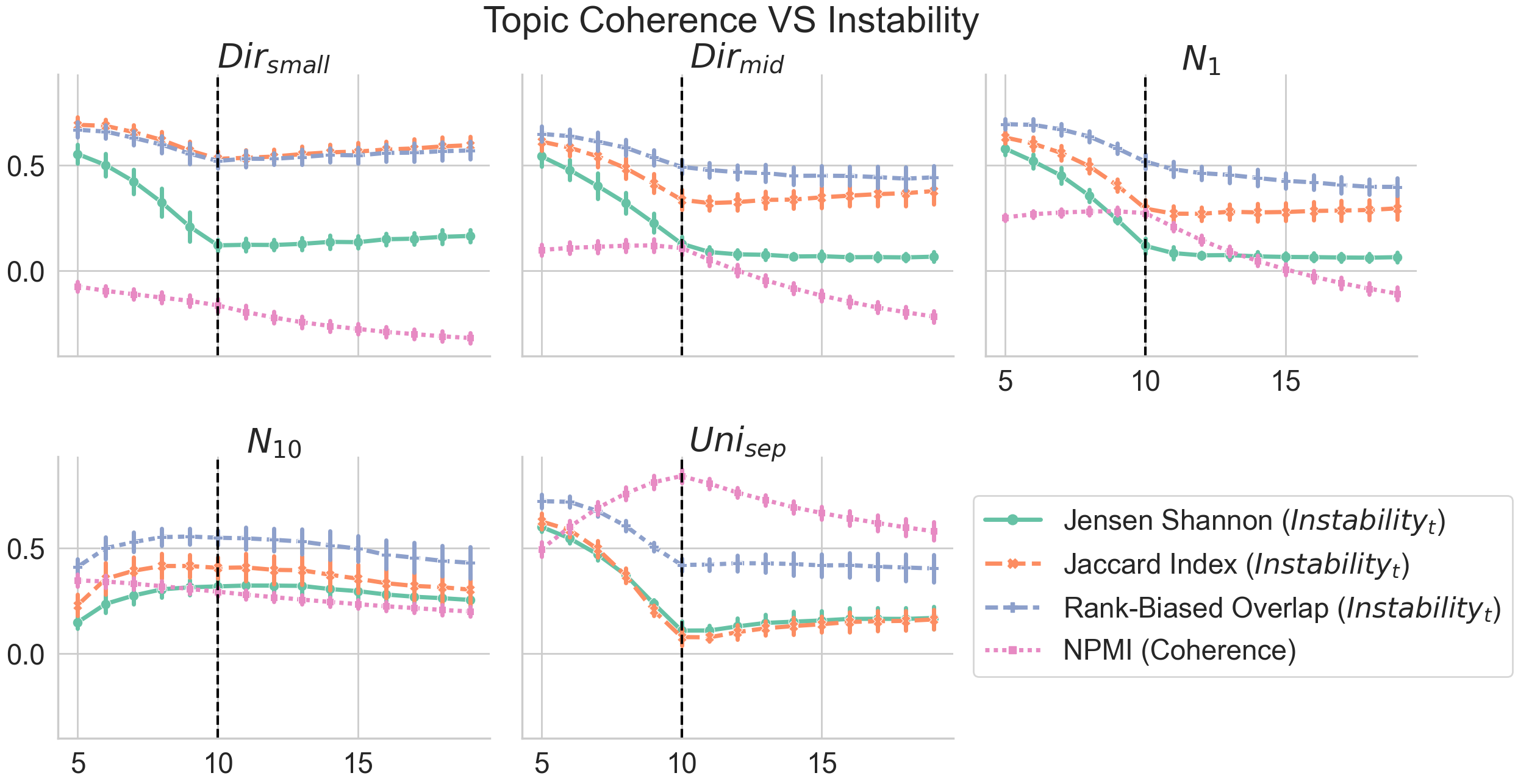}
    \caption{The three \textit{Instability} measures and NPMI coherence across topic counts for each simulated distribution. Points and error bars show the means and standard deviations across 50 simulated corpora. The dashed line indicates the true topic count \(K=10\). Lower \textit{Instability} indicates greater repeatability, while higher NPMI indicates greater coherence. The measures share an axis for visual comparison only.}
    \label{fig:coherence}
\end{figure}

% \begin{figure}[ht]
%     \centering
%     \includegraphics[width = \textwidth]{../figure/instabilty_larger_range.png}
%     \caption{The plot shows the instability measure as the number of topics increases, where \(K = 10\) is the true number of topics. The different colours indicate different distributions used for generation, and each plot represents a different similarity measure. For most of the distributions, we see that it dips or plateaus around \(K=10\), indicating the measure is most stable at that point. Looking at \(N_{10}\), the measurements do not perform well.}
%     \label{fig:topics_mean_dist}
% \end{figure}

Figure \ref{fig:coherence} shows how the \textit{Instability} measure, \(Instability_t\), performs for each similarity measure used. For Jensen-Shannon Similarity, we see that all distributions dip at \(K=10\), the true number of topics, except for the Normal Distribution with standard deviation 10 (\(N_{10}\)). The values are relatively the same across the distributions. Similarly, for Jaccard Similarity, we see the \textit{Instability} values plateau at \(K=10\), except for the normal distribution \(N_{10}\), and the steepest change is with the Uniform Separable distribution (\(Uni_{sep}\)). Rank-Biased Overlap uses both the top terms and their ordering. We see that \(Uni_{sep}\) drops at \(K=10\), then the \textit{Instability} values plateau. This pattern can also be seen with the Dirichlet Distribution with a smaller parameter (\(Dir_{small}\)); however, it is harder to distinguish from the other distributions.

We can observe that when comparing distributions, \textit{i.e.}, the Jensen-Shannon Similarity, the different distributions used make little difference in the \textit{Instability} measure. For Jaccard Similarity, excluding \(N_{10}\), where we only consider the top 10 words of each topic, the most stable distribution is \(Uni_{sep}\), followed by \(N_{1}\), \(Dir_{mid}\), and finally \(Dir_{small}\). When using the ranked top words, \textit{i.e.}, Rank-Biased Overlap (RBO), we see that the most stable distribution is once again \(Uni_{sep}\), followed by \(Dir_{mid}\), while \(N_{1}\) and \(Dir_{small}\) are tied.

The Instability measures generally provide a signal near the true number of topics, although the strength of this signal depends on the distribution and similarity measure. Jensen-Shannon Similarity shows relatively consistent behaviour across the different distributions, suggesting that it is less sensitive to changes in the data-generating process. Jaccard Similarity, which compares the top words of each topic, is more sensitive to the separation between the topic distributions. This is most noticeable for \(Uni_{sep}\), where the change at \(K=10\) is more pronounced, but also for \(N_{10}\), where the distinction is less clear. Rank-Biased Overlap, which incorporates both the top terms and their ordering, produces similar behaviour to Jaccard Similarity but with less separation between some of the distributions. These results show that the measured \textit{Instability} depends on both the similarity measure used and the underlying topic distributions.

\subsection{Comparison with Topic Coherence}

To compare the proposed Instability measure with an established topic-quality measure, NPMI coherence was calculated across the same range of topic numbers for each simulated corpus. NPMI coherence was calculated using a sliding reference window of 10 tokens. Figure 1 compares coherence with the three Instability measures for each generating distribution. The vertical dashed line indicates the true number of generating topics \(K=10\). The \textit{Instability} values shown across the complete range of \(K\) compare repeated LDA outputs and therefore measure internal repeatability. Recovery relative to the ground-truth distributions is examined separately in Section \ref{sec:within}.

We can see that coherence behaves differently across the simulated corpora. For the Uniform Separable distribution, NPMI coherence reaches its maximum at \(K=10\), showing agreement between coherence and the Instability measures when the underlying topics are clearly separated. The \textit{Instability} values also decrease sharply near \(K=10\), indicating that repeated LDA fits become more similar near the true number of topics.

However, a different pattern can be observed for the Dirichlet and Normal distributions. For \(Dir_{small}\), coherence generally decreases as the number of topics increases. For \(Dir_{mid}\) and \(N_1\), coherence increases slightly at the smaller topic counts before decreasing around and beyond \(K=10\). Coherence for \(N_{10}\) changes less around \(K=10\) but also decreases at the larger topic counts. Therefore, for these distributions, coherence generally favours a smaller number of topics rather than a number larger than the true value.

Across the simulated distributions, NPMI agreed most clearly with the Instability measures for the Uniform Separable distribution. For the more overlapping distributions, NPMI generally favoured smaller topic counts, whereas the Instability measures produced a minimum, plateau, or noticeable change near \(K=10\). This pattern was weakest for \(N_{10}\).

% The differences between coherence and Instability become more noticeable as the topic-word distributions become more overlapping. A smaller value of \(K\) can combine terms into broader topics with strongly associated highest-ranked words, resulting in higher coherence. However, this does not mean that the fitted model represents the known number of generating topics. In comparison, the Instability measures generally show a minimum, plateau, or noticeable change near \(K=10\), although this pattern is weaker for \(N_{10}\).

% These results show that coherence and Instability capture different properties of a topic model. Coherence measures the association between the highest-ranked terms within the fitted topics, whereas the Instability values in Figure 1 measure the similarity of the topics across repeated LDA fits. Agreement between the measures for the Uniform Separable distribution provides stronger evidence for \(K=10\). When the topic distributions overlap, the disagreement between the measures provides useful information about the difference between semantic coherence and repeatability.

% Overall, coherence and the Instability measures agree most clearly when the topic structure is well separated. For the more overlapping distributions, coherence tends to favour smaller topic counts, while the \textit{Instability} measures generally provide a signal near the true value of \(K=10\). Using the measures together therefore provides a more complete assessment of topic-model performance than using coherence alone.

\subsection{Recovery and Repeatability}\label{sec:within}

\begin{figure*}[ht]
    \centering
    \includegraphics[width = \textwidth]{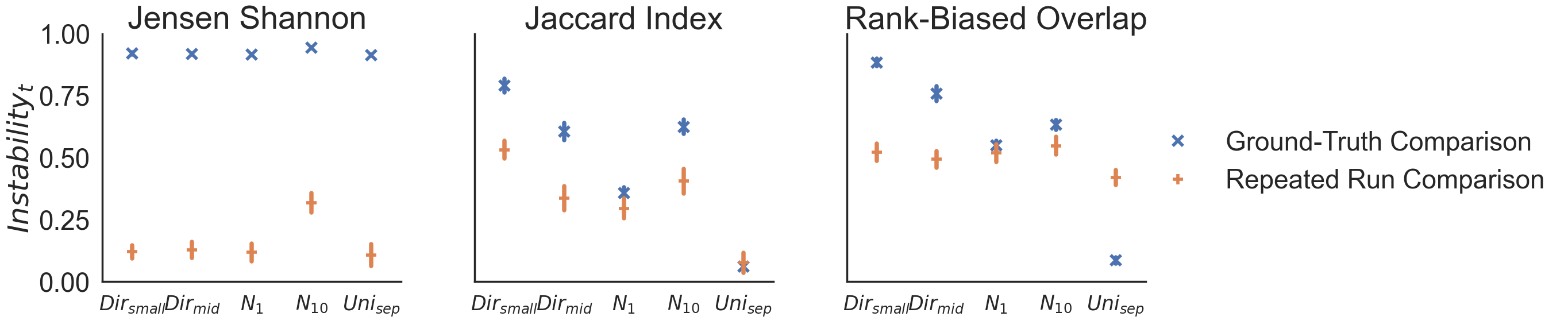}
    \caption{\textit{Instability} values for the five simulated topic-word distributions. Blue crosses compare fitted topics with the ground truth, while orange plus signs compare repeated LDA fits. Points show the mean across 50 simulated corpora, and error bars show the standard deviation of the corpus-specific \textit{Instability} values. Lower values indicate greater similarity and repeatability. LDA is generally more repeatable than accurate, except for the Uniform Separable distribution under Jaccard Similarity and Rank-Biased Overlap.}
    \label{fig:between_within_mean}
\end{figure*}

Figure \ref{fig:between_within_mean} shows the \textit{Instability} measure applied to ground truth data and repeated run and how they compare. Here, the blue values represent LDA outputs compared to the `true' distributions, and the orange values represent comparisons between repeated LDA runs. Comparing to the `true' distributions determines the recovery accuracy of LDA. The comparisons between repeated runs determine the consistency of the LDA outputs.

From Figure \ref{fig:between_within_mean}, we can observe that the repeated-run Instability values are generally lower than the ground-truth Instability values. This shows that LDA is generally more consistent across repeated runs than it is accurate at recovering the `true' topic distributions. However, the size of this difference depends on the similarity measure and the distribution used to generate the corpus.

For Jensen–Shannon Similarity, the repeated-run Instability values are much lower than the ground-truth values for all five distributions. This difference is also observed for the Uniform Separable distribution. The repeated-run value is higher for \(N_{10}\) than for the other distributions, but it is still substantially lower than the comparison with the `true' distribution. This shows that the complete topic-word distributions produced by LDA are more similar to each other than to the generating distributions.

For Jaccard Similarity, the repeated-run \textit{Instability} values are lower for \(Dir_{small}\), \(Dir_{mid}\), \(N_1\), and \(N_{10}\). The difference is smaller for \(N_1\) and larger for the two Dirichlet distributions and \(N_{10}\). For the Uniform Separable distribution, both \textit{Instability} values are close to zero, and the comparison with the `true' distribution performs slightly better. Therefore, when the topics are completely separated, LDA accurately recovers the sets containing the 10 highest-ranked terms.

Rank-Biased Overlap shows a similar pattern. The repeated-run Instability values are lower for the two Dirichlet distributions and \(N_{10}\), while the two values are relatively similar for \(N_1\). For the Uniform Separable distribution, the comparison with the ‘true’ distribution performs better than the repeated-run comparison. This shows that LDA can recover both the highest-ranked terms and their ordering when the topics are clearly separated.

We can also observe that Jensen–Shannon Similarity shows the largest differences between the ground-truth and repeated-run comparisons. This measure compares the complete topic-word distributions and therefore captures changes across the full vocabulary. Jaccard Similarity and Rank-Biased Overlap only compare the 10 highest-ranked terms and show better recovery for the Uniform Separable distribution.

Across most generating distributions, repeated-run Instability was lower than ground-truth Instability. The exception was the Uniform Separable distribution under Jaccard Similarity and Rank-Biased Overlap, where ground-truth recovery was similar to or better than repeatability.

\subsection{20 Newsgroups Data}

% \begin{wrapfigure}{r}{0.5\textwidth}
%    \centering
%     \includegraphics[width = \textwidth]{../figure/news_stability_t.png}
%     \caption{The stability distance measure applied 20Newsgroup dataset over a range of the number of topics. We see that the measures dip around \(K=20\), which is the true number of topics.}
%     \label{fig:newsgroup}
% \end{wrapfigure}

To compare the proposed \textit{Instability} framework with an established topic-quality metric, NPMI topic coherence was computed for the same range of topic numbers on the 20 Newsgroups corpus. Figure 3 shows the behaviour of the three \textit{Instability} measures alongside NPMI coherence. The corpus contains 20 labelled newsgroup categories that serve as a reference point. However, because the generating topic distributions are unknown, the \textit{Instability} measures can only be used to assess consistency across repeated fits, not recovery accuracy.

\begin{figure}[ht]
    \centering
    \includegraphics[width=\textwidth]{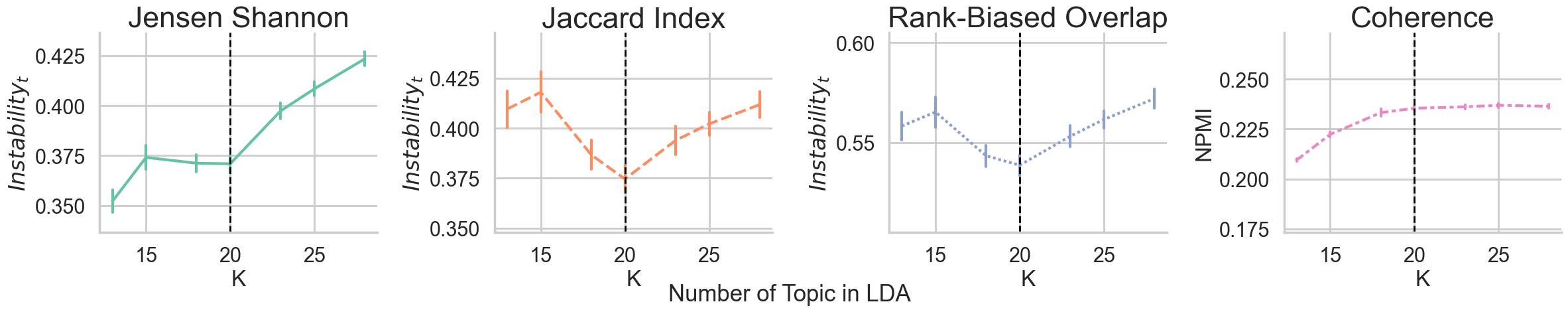}
    \caption{The three \textit{Instability} measures and NPMI coherence for the 20 Newsgroups dataset. The dashed line indicates the 20 labelled categories. For the \textit{Instability} measures, error bars show the standard deviation of the \textit{Instability} value. For NPMI, error bars show the standard deviation of the coherence scores across the 50 fits.}
    \label{fig:newsgroup}
\end{figure}

We can see that the \textit{Instability} measures based on Jaccard Similarity and Rank-Biased Overlap show clear minimum values at approximately \(K=20\). This suggests that repeated LDA fits produce their most similar topic-word distributions near the number of labelled categories when the topics are compared using their highest-ranked terms.

However, the pattern is different for Jensen–Shannon Similarity. Jensen–Shannon \textit{Instability} is lowest at the smallest value of \(K\) considered, increases initially, and changes only slightly around \(K=20\) before increasing again at larger values of \(K\). Therefore, it does not show the same minimum near 20 topics as the Jaccard and Rank-Biased Overlap measures. This indicates that the observed level of \textit{Instability} depends on whether the complete topic-word distributions or only the highest-ranked terms are compared.

NPMI coherence also shows a different pattern. Coherence generally increases as the number of topics increases and reaches its highest values above \(K=20\), where it begins to level off. Jaccard and Rank-Biased Overlap Instability reached their minimum values near \(K=20\). Jensen–Shannon Instability was lowest at the smallest topic count considered, while NPMI increased beyond \(K=20\) before levelling off.

% To evaluate the stability measure on a real dataset, we applied it to the 20 Newsgroups dataset, as shown in Figure \ref{fig:newsgroup}. We can see that Jaccard Similarity and Rank-Based Overlap dip at \(K = 20\), indicating that LDA is most stable around \(K = 20\). However, it is less clear when using the Jensen-Shannon Similarity based only on the distribution of topics. The stability measure plateaus around \(K = 20\) and then shoots up thereafter. This could be because the true number of topics isn't well known, as some topics overlap in the dataset. 

\section{Discussion and Conclusion}

Repeatability is not the same as recovery, \textit{i.e.}, as we have shown that topic model can repeatedly return similar topics without accurately recovering the `true' underlying distribution. The ground-truth comparisons showed that LDA was generally more consistent across repeated fits than it was accurate relative to the generating topic-word distributions. This means that repeatability provides evidence that the fitting procedure can yield a similar solution, but it does not, by itself, show that the solution is correct. Repeatability should therefore be treated as one part of topic-model validation rather than as evidence of recovery.

Repeated fitting remains important because it makes uncertainty caused by the fitting procedure visible. In this study, each of the 50 independently generated corpora was fitted 50 times using different fixed random seeds, with no fits excluded. The results therefore capture variation caused by both corpus sampling and stochastic model fitting, rather than relying on a single corpus or a selected subset of model outputs. If only one fit is reported, it is not possible to determine whether the resulting topics are representative of the model’s general behaviour or specific to the selected random seed. Repeated fitting provides a practical way to make this distinction.

The results also show that \textit{Instability} can provide useful evidence when selecting the number of topics. Across the simulated corpora, the \textit{Instability} measures generally produced a minimum, plateau, or noticeable change near \(K=10\), which was the known number of generating topics. The signal was clearest when the generating topics were well separated and became less distinct as the topics overlapped. This means that \textit{Instability} can identify topic counts at which the fitted representation becomes comparatively repeatable, while also making the uncertainty in this choice visible. However, the minimum Instability value should not automatically be interpreted as the true number of topics. It identifies a comparatively repeatable solution, and the meaning of that solution still depends on the corpus's structure and the purpose of the analysis.

Coherence provides different evidence about the appropriate number of topics. NPMI agreed with the \textit{Instability} measures for the Uniform Separable distribution, where the generating topics were clearly separated. For the more overlapping Dirichlet and Normal distributions, NPMI generally favoured smaller topic counts. A smaller number of topics can combine related terms into broader groups, producing high coherence even when the model does not recover the known number of generating topics. Therefore, high coherence can indicate that the highest-ranked words are strongly related, but it does not necessarily indicate that the fitted model has recovered the generating structure.

This distinction has practical implications because the most appropriate number of topics can depend on how the topic model output will be used. A smaller set of broad, coherent topics may be useful when the aim is to provide a general summary of a large corpus. A larger number of more specific topics may be more useful when the aim is to distinguish between closely related themes. Combining several generated topics into one broader topic may improve coherence while removing distinctions that are important to the research question. Topic-count selection should therefore consider both the numerical measures and the level of detail required from the analysis.

The part of the topic representation being compared is also important. Jaccard Similarity and Rank-Biased Overlap evaluate the highest-probability terms, which are usually the terms presented when a topic is interpreted or labelled. Low Instability under these measures indicates that similar representative terms are returned across repeated fits. Rank-Biased Overlap provides further information by considering whether these terms also occur in a similar order. These measures are therefore particularly relevant when the fitted topics are primarily used to summarise a corpus or provide qualitative labels.

Jensen–Shannon Similarity evaluates the complete topic-word probability distributions and can identify differences that are not visible among the highest-ranked terms. The Uniform Separable results provide evidence for this distinction. Jaccard Similarity and Rank-Biased Overlap showed that LDA recovered the representative terms and their ordering, while Jensen–Shannon Similarity showed that the complete probability distributions were not recovered to the same extent. This meant that the model identified the words that represented the topics more accurately than it estimated the probabilities of all of them. This difference is important when the complete topic-word distributions are used to compare topics or support later quantitative analysis.

Disagreement between the similarity measures is therefore useful rather than contradictory.  Agreement between the three Instability measures provides stronger evidence that both the representative terms and the wider distributions are repeatable. When the measures disagree, their differences show where the variation occurs. Low Jaccard and Rank-Biased Overlap, with higher Jensen–Shannon Instability, indicate that the main interpretation of the topics is consistent, but that the probabilities across the wider vocabulary are less consistent. Using the measures together provides more information than selecting one similarity measure as correct in every setting.

The 20 Newsgroups results show how these measures can be interpreted when the generating distributions are unknown. Jaccard and Rank-Biased Overlap Instability were lowest near the 20 labelled newsgroup categories, showing that the highest-ranked terms and their ordering were most repeatable near this reference point. Jensen–Shannon Instability and NPMI did not show the same pattern. This indicates that the labelled categories have a meaningful relationship with the most repeatable representative terms, but not necessarily with the complete probability distributions or the most coherent topics.

The labelled categories provide an external reference point for the latent topic structure. A newsgroup category may contain several topics, while related topics may occur across multiple categories. The number of categories, therefore, provides an external reference point rather than a known ground truth. Even so, the agreement between the two top-term measures and this reference point is a useful result. It shows that the framework can identify an interpretable pattern in real-world data while remaining clear about the difference between repeatability and recovery.

More generally, real-world corpora do not provide direct access to the distributions that generated their topics. This makes it difficult to determine whether a fitted model has recovered an underlying structure or produced a single useful representation among several possible ones. Stability can indicate whether the result is sensitive to repeated fitting, but it cannot resolve this question on its own. Topic-model conclusions should therefore be supported using multiple forms of evidence, including repeated fitting, coherence, external information where available, and direct examination of the fitted topics.

This is particularly important when topic-model outputs support decisions in high-stakes decision-making settings. Similar topic labels across repeated fits may appear to provide strong validation, but this study shows that the full distributions can remain different even when the highest-ranked terms are stable. The strength of a conclusion should therefore reflect the part of the model output that has been shown to be repeatable. Conclusions based on representative terms can be supported by Jaccard Similarity and Rank-Biased Overlap, whereas conclusions that rely on complete topic-word probabilities require evidence from distribution-based comparisons, such as Jensen–Shannon Similarity.

The controlled corpus-generation process is a strength of the current study because it provides a clear setting in which recovery can be measured directly. The generating distributions were known, the level of topic separation could be varied, and variation from corpus sampling could be separated from variation in model fitting. Although these simulated corpora follow the LDA generative process and use relatively simple predefined distributions, NPMI coherence still did not consistently identify the known number of topics. This provides a useful warning for more complex real-world corpora, where the generating structure and level of topic separation are unknown. If coherence can be difficult to interpret under controlled conditions, greater care may be required when it is applied to more complex data.

Starting with one LDA implementation and a fixed fitting procedure also provided a consistent basis for comparing the measures. The experiments used the tomotopy Gibbs-sampling implementation, comparisons of the estimated topic-word distributions, the 10 highest-probability terms for Jaccard Similarity and Rank-Biased Overlap, and a greedy topic-matching procedure based on Jensen–Shannon Similarity. Keeping these choices fixed allowed the effects of the generating distributions and evaluation measures to be examined clearly. These choices also identify direct opportunities to extend the framework rather than limit its wider use.

Future work can apply the same framework to alternative inference procedures, topic-matching methods, persistence parameters, and numbers of highest-ranked terms. Comparing matching procedures based on different similarity measures would show whether the alignment method affects the resulting \textit{Instability} values. The framework can also be extended to document-topic distributions, other topic models, and simulated corpora with more complex linguistic structure. Since the framework separates repeatability from recovery, it can provide a common basis for comparing these extensions under controlled conditions before they are applied to real-world data.

This paper contributes a controlled process for generating corpora with known topic-word distributions and an Instability framework for comparing recovery with consistency across repeated topic-model fits. The results show that repeatability, recovery, and coherence capture distinct aspects of topic-model performance. A model can repeatedly return similar topics without accurately recovering the distributions that generated the data. Repeatability should therefore not be treated on its own as evidence that the fitted topics are correct.

\paragraph{Competing Interests}
The authors declare that they have no known competing financial interests or personal relationships that could have appeared to influence the work reported in this paper.

\paragraph{Funding Statement}
This research received no specific grant from any funding agency, commercial, or not-for-profit
sectors.

\paragraph{Data Availability Statement}
Replication code, simulation scripts, and derived results are available upon request from the corresponding author.

\paragraph{Ethics Statement}
 Ethical approval was not required.

\paragraph{Acknowledgements}

During manuscript preparation, the authors used Grammarly web application for sentence-level language and clarity editing. Grammarly was not used to generate research ideas, analyses, data, citations or conclusions. All suggestions were reviewed and revised by the authors, who take full responsibility for the final text.

\printbibliography

@inproceedings{doi2024topic,
  title={Topic modeling for short texts with large language models},
  author={Doi, Tomoki and Isonuma, Masaru and Yanaka, Hitomi},
  booktitle={Proceedings of the 62nd Annual Meeting of the Association for Computational Linguistics (Volume 4: Student Research Workshop)},
  pages={21--33},
  year={2024}
}

@article{vayansky2020review,
  title={A review of topic modeling methods},
  author={Vayansky, Ike and Kumar, Sathish AP},
  journal={Information Systems},
  volume={94},
  pages={101582},
  year={2020},
  publisher={Elsevier}
}

@article{ballester2022robustness,
  title={Robustness, replicability and scalability in topic modelling},
  author={Ballester, Omar and Penner, Orion},
  journal={Journal of Informetrics},
  volume={16},
  number={1},
  pages={101224},
  year={2022},
  publisher={Elsevier}
}

@article{abdelrazek2023topic,
  title={Topic modeling algorithms and applications: A survey},
  author={Abdelrazek, Aly and Eid, Yomna and Gawish, Eman and Medhat, Walaa and Hassan, Ahmed},
  journal={Information Systems},
  volume={112},
  pages={102131},
  year={2023},
  publisher={Elsevier}
}

@article{jelodar2019latent,
  title={Latent Dirichlet allocation (LDA) and topic modeling: models, applications, a survey},
  author={Jelodar, Hamed and Wang, Yongli and Yuan, Chi and Feng, Xia and Jiang, Xiahui and Li, Yanchao and Zhao, Liang},
  journal={Multimedia tools and applications},
  volume={78},
  pages={15169--15211},
  year={2019},
  publisher={Springer}
}

@misc{20newsgroups,
  author       = {Ken Lang},
  title        = {20 Newsgroups Dataset},
  year         = {1995},
  howpublished = {\url{http://qwone.com/~jason/20Newsgroups/}},
  urldate = {2025-04-07}
}

@incollection{blei2009topic,
  title={Topic models},
  author={Blei, David M and Lafferty, John D},
  booktitle={Text mining},
  pages={101--124},
  year={2009},
  publisher={Chapman and Hall/CRC}
}

@article{hambarde2023information,
  title={Information retrieval: recent advances and beyond},
  author={Hambarde, Kailash A and Proenca, Hugo},
  journal={IEEE Access},
  volume={11},
  pages={76581--76604},
  year={2023},
  publisher={IEEE}
}

@inproceedings{onah2022data,
  title={A data-driven latent semantic analysis for automatic text summarization using lda topic modelling},
  author={Onah, Daniel FO and Pang, Elaine LL and El-Haj, Mahmoud},
  booktitle={2022 IEEE International Conference on Big Data (Big Data)},
  pages={2771--2780},
  year={2022},
  organization={IEEE}
}

@article{murakami2017corpus,
  title={‘What is this corpus about?’: using topic modelling to explore a specialised corpus},
  author={Murakami, Akira and Thompson, Paul and Hunston, Susan and Vajn, Dominik},
  journal={Corpora},
  volume={12},
  number={2},
  pages={243--277},
  year={2017},
  publisher={Edinburgh University Press The Tun-Holyrood Road, 12 (2f) Jackson's Entry~…}
}

@inproceedings{lau2012line,
  title={On-line trend analysis with topic models:\# twitter trends detection topic model online},
  author={Lau, Jey Han and Collier, Nigel and Baldwin, Timothy},
  booktitle={Proceedings of COLING 2012},
  pages={1519--1534},
  year={2012}
}

@article{corti2022social,
  title={Social media analysis of Twitter tweets related to ASD in 2019--2020, with particular attention to COVID-19: topic modelling and sentiment analysis},
  author={Corti, Luca and Zanetti, Michele and Tricella, Giovanni and Bonati, Maurizio},
  journal={Journal of big data},
  volume={9},
  number={1},
  pages={113},
  year={2022},
  publisher={Springer}
}

@inproceedings{howes2013investigating,
  booktitle={Investigating topic modelling for therapy dialogue analysis},
  title = {Investigating topic modelling for therapy dialogue analysis},
  author={Howes, Christine and Purver, Matthew and McCabe, Rose},
  year={2013},
  organization={Association for Computational Linguistics}
}

@inproceedings{mantylaMeasuringLDATopic2018,
  title = {Measuring {{LDA}} Topic Stability from Clusters of Replicated Runs},
  booktitle = {Proceedings of the 12th {{ACM}}/{{IEEE International Symposium}} on {{Empirical Software Engineering}} and {{Measurement}}},
  year = {2018},
  author = {Mantyla, Mika V. and Claes, Maelick and Farooq, Umar},
  date = {2018-10-11},
  series = {{{ESEM}} '18},
  pages = {1--4},
  publisher = {Association for Computing Machinery},
  location = {New York, NY, USA},
  doi = {10.1145/3239235.3267435},
  urldate = {2022-11-16},
  isbn = {978-1-4503-5823-1}
}

@incollection{greeneHowManyTopics2014,
  title = {How {{Many Topics}}? {{Stability Analysis}} for {{Topic Models}}},
  year = {2014},
  shorttitle = {How {{Many Topics}}?},
  booktitle = {Machine {{Learning}} and {{Knowledge Discovery}} in {{Databases}}},
  author = {Greene, Derek and O’Callaghan, Derek and Cunningham, Pádraig},
  editor = {Calders, Toon and Esposito, Floriana and Hüllermeier, Eyke and Meo, Rosa},
  date = {2014},
  volume = {8724},
  pages = {498--513},
  publisher = {Springer Berlin Heidelberg},
  location = {Berlin, Heidelberg},
  doi = {10.1007/978-3-662-44848-9_32},
  urldate = {2022-10-14},
  isbn = {978-3-662-44847-2 978-3-662-44848-9},
  langid = {english}
}

@article{endresNewMetricProbability2003,
  title = {A New Metric for Probability Distributions},
  author = {Endres, D.M. and Schindelin, J.E.},
  year = {2003},
  month = jul,
  journal = {IEEE Transactions on Information Theory},
  volume = {49},
  number = {7},
  pages = {1858--1860},
  issn = {0018-9448},
  doi = {10.1109/TIT.2003.813506},
  urldate = {2025-02-24},
  copyright = {https://ieeexplore.ieee.org/Xplorehelp/downloads/license-information/IEEE.html},
  langid = {english}
}

@article{rbo,
author = {Webber, William and Moffat, Alistair and Zobel, Justin},
title = {A similarity measure for indefinite rankings},
year = {2010},
issue_date = {November 2010},
publisher = {Association for Computing Machinery},
address = {New York, NY, USA},
volume = {28},
number = {4},
issn = {1046-8188},
doi = {10.1145/1852102.1852106},
journal = {ACM Trans. Inf. Syst.},
month = nov,
articleno = {20},
numpages = {38}
}

@inproceedings{taylorSimLDAToolTopic2023,
  title = {{{SimLDA}}: {{A Tool}} for~{{Topic Model Evaluation}}},
  shorttitle = {{{SimLDA}}},
  booktitle = {Proceedings of the {{Future Technologies Conference}} ({{FTC}}) 2022, {{Volume}} 3},
  author = {Taylor, Rebecca M. C. and du Preez, Johan A.},
  editor = {Arai, Kohei},
  year = {2023},
  series = {Lecture {{Notes}} in {{Networks}} and {{Systems}}},
  pages = {534--554},
  publisher = {{Springer International Publishing}},
  location = {{Cham}},
  doi = {10.1007/978-3-031-18344-7_38},
  isbn = {978-3-031-18344-7},
  langid = {english}
}

@article{weisserPseudodocumentSimulationComparing2023,
  title = {Pseudo-Document Simulation for Comparing {{LDA}}, {{GSDMM}} and {{GPM}} Topic Models on Short and Sparse Text Using {{Twitter}} Data},
  author = {Weisser, Christoph and Gerloff, Christoph and Thielmann, Anton and Python, Andre and Reuter, Arik and Kneib, Thomas and Säfken, Benjamin},
  date = {2023-06-01},
  year = {2023},
  journal = {Computational Statistics},
  shortjournal = {Comput Stat},
  volume = {38},
  number = {2},
  pages = {647--674},
  issn = {1613-9658},
  doi = {10.1007/s00180-022-01246-z},
  urldate = {2024-01-28},
  langid = {english}
}

@inproceedings{roder2015,
author = {R\"{o}der, Michael and Both, Andreas and Hinneburg, Alexander},
title = {Exploring the Space of Topic Coherence Measures},
year = {2015},
isbn = {9781450333177},
publisher = {Association for Computing Machinery},
address = {New York, NY, USA},
doi = {10.1145/2684822.2685324},
booktitle = {Proceedings of the Eighth ACM International Conference on Web Search and Data Mining},
pages = {399–408},
numpages = {10},
location = {Shanghai, China},
series = {WSDM '15}
}

@article{belfordStabilityTopicModeling2018a,
	title = {Stability of topic modeling via matrix factorization},
	volume = {91},
    year = {2018},
	issn = {0957-4174},
	doi = {10.1016/j.eswa.2017.08.047},
	pages = {159--169},
	journal = {Expert Systems with Applications},
	shortjournal = {Expert Systems with Applications},
	author = {Belford, Mark and Mac Namee, Brian and Greene, Derek},
	urldate = {2024-01-28},
	date = {2018-01-01},
}

\end{document}